\documentclass[11pt,a4paper]{article}
\usepackage[hyperref]{acl2021}
\usepackage{times}
\usepackage{latexsym}

\usepackage{microtype}

\usepackage{times}
\usepackage{latexsym}
\usepackage{amsthm,amsmath,amssymb}
\usepackage{multirow}
\usepackage{array}
\usepackage{mathrsfs}
\usepackage{amsmath}
\usepackage{amsfonts}
\usepackage{graphicx}
\usepackage{booktabs}
\usepackage{algorithm}
\usepackage{algorithmicx}
\usepackage{algpseudocode}
\usepackage{amsfonts}
\usepackage{yhmath}
\usepackage{url}
\usepackage{tikz}
\usepackage{pgfplots}
\usepackage{makecell}
\usepackage{array}
\usepackage{tabularx}

\usepackage{subcaption}
\usetikzlibrary{shapes, positioning, decorations.pathreplacing, shapes.multipart}

\usepackage{scrextend}

\aclfinalcopy % Uncomment this line for the final submission
\title{A Unified Generative Framework for Various NER Subtasks}

\author{Hang Yan\textsuperscript{1}, Tao Gui\textsuperscript{2}, Junqi Dai\textsuperscript{1}, Qipeng Guo\textsuperscript{1}, Zheng Zhang\textsuperscript{3}, Xipeng Qiu\textsuperscript{1,4}\thanks{\ \  Corresponding author.}\\
  \textsuperscript{1}Shanghai Key Laboratory of Intelligent Information Processing, Fudan University \\
  \textsuperscript{1}School of Computer Science, Fudan University \\
  \textsuperscript{2}Institute of Modern Languages and Linguistics, Fudan University \\
  \textsuperscript{3}New York University \\
  \textsuperscript{4}Pazhou Lab, Guangzhou, China \\
  \texttt{\{hyan19,tgui16,jqdai19,qpguo16,xpqiu\}@fudan.edu.cn}\\
  \texttt{zz@nyu.edu}\\}

\date{}

\begin{document}
\maketitle
\begin{abstract}
  Named Entity Recognition (NER) is the task of identifying spans that represent entities in sentences. Whether the entity spans are nested or discontinuous, the NER task can be categorized into the flat NER, nested NER, and discontinuous NER subtasks. These subtasks have been mainly solved by the token-level sequence labelling or span-level classification. However, these solutions can hardly tackle the three kinds of NER subtasks concurrently. To that end, we propose to formulate the NER subtasks as an entity span sequence generation task, which can be solved by a unified sequence-to-sequence (Seq2Seq) framework. Based on our unified framework, we can leverage the pre-trained Seq2Seq model to solve all three kinds of NER subtasks without the special design of the tagging schema or ways to enumerate spans. We exploit three types of entity representations to linearize entities into a sequence. Our proposed framework is easy-to-implement and achieves state-of-the-art (SoTA) or near SoTA performance on eight English NER datasets, including two flat NER datasets, three nested NER datasets, and three discontinuous NER datasets \footnote{Code is available at \url{https://github.com/yhcc/BARTNER}.}.
\end{abstract}
% NER是一个识别文本中的entity span的任务。根据enentity是否是连续的span或者是否具有嵌套关系的span，ner被分成了三种不同的任务。

\section{Introduction} \label{sec:intro}

% 研究任务是什么, 首先简要引入一下NER
Named entity recognition (NER) has been a fundamental task of Natural Language Processing (NLP), and three kinds of NER subtasks have been recognized in previous work \citep{DBLP:conf/conll/SangM03,DBLP:conf/clef/PradhanESMCVSCS13,DBLP:conf/lrec/DoddingtonMPRSW04,DBLP:conf/ismb/KimOTT03,DBLP:journals/jbi/KarimiMKW15}, including flat NER, nested NER, and discontinuous NER.
As shown in Figure~\ref{fig:example}, the nested NER contains overlapping entities, and the entity in the discontinuous NER may contain several nonadjacent spans.
% As depicted in Fig.\ref{fig:example1}, the same string ``Louis Vuitton'' can appear in different entities. Therefore, the predicted span must conatin the entity's start and end indexes to disambiguate. Moreover, one word can be contained by more than one entities as the examples shown the nested NER and discontinuous NER tasks. Specially, in the discontinuous NER task, one entity may contain discontinuous tokens as Fig.\ref{fig:example3} shows.
% 这里需要补充一下各种NER的基本情况. Flat NER, nested NER

\begin{figure}[!h]
  \centering
  \includegraphics[width=0.95\columnwidth]{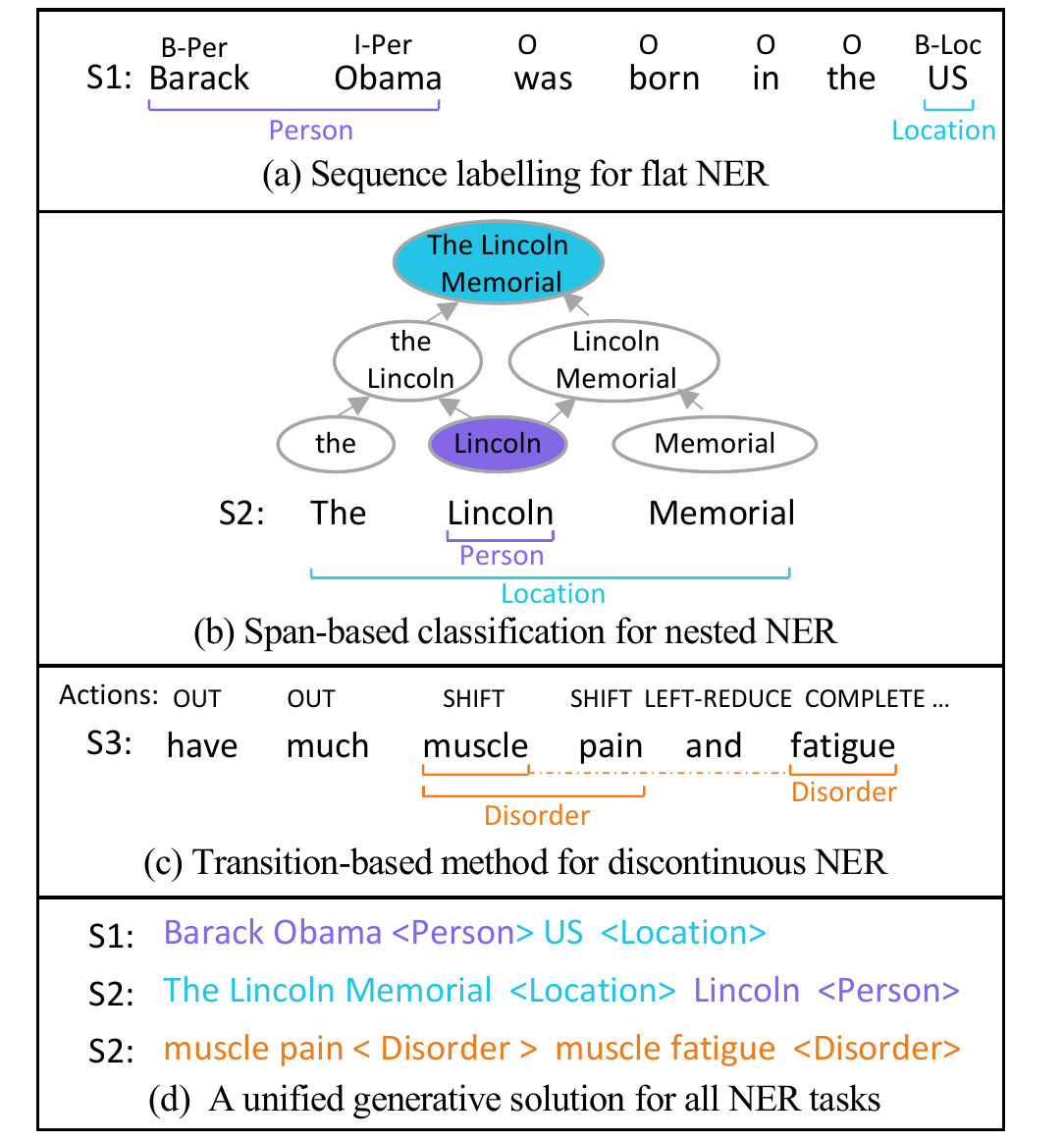} \label{fig:example}
  \vspace{-0.3em}
  \caption{Examples of three kinds of NER subtasks. (a) - (c) illustrate flat NER, nested NER, discontinuous NER, and their corresponding mainstream solutions respectively. (d) Our proposed generative solution to solve all NER subtasks in a unified way.} \label{fig:example}
\end{figure}

% 已有的方法. 这里还是想要围绕这序列标注的范式来写，
The sequence labelling formulation, which will assign a tag to each token in the sentence, has been widely used in the flat NER field \citep{DBLP:conf/conll/McCallum003,DBLP:journals/jmlr/CollobertWBKKK11,DBLP:journals/corr/HuangXY15,DBLP:journals/tacl/ChiuN16,DBLP:conf/naacl/LampleBSKD16,DBLP:conf/acl/StrakovaSH19,DBLP:journals/corr/abs-1911-04474,DBLP:conf/acl/LiYQH20}. Inspired by sequence labelling's success in the flat NER subtask, \citet{DBLP:conf/semweb/Metke-JimenezK16,DBLP:conf/emnlp/MuisL17} tried to formulate the nested and discontinuous NER into the sequence labelling problem. For the nested and discontinuous NER subtasks, instead of assigning labels to each token directly, \citet{DBLP:conf/acl/XuJW17,DBLP:conf/emnlp/WangL19,DBLP:conf/acl/YuBP20,DBLP:conf/acl/LiFMHWL20} tried to enumerate all possible spans and conduct the span-level classification. Another way to efficiently represent spans is to use the hypergraph \citep{DBLP:conf/emnlp/LuR15,DBLP:conf/naacl/KatiyarC18,DBLP:conf/emnlp/WangL18,DBLP:conf/emnlp/MuisL16}.

% 把过去人们怎么做NER task的情况说明一下

% 面临的挑战
Although the sequence labelling formulation has dramatically advanced the NER task, it has to design different tagging schemas to fit various NER subtasks. One tagging schema can hardly fit for all three NER subtasks\footnote{Attempts made for discontinuous constituent parsing may tackle three NER subtasks in one tagging schema \cite{DBLP:conf/emnlp/VilaresG20}.} \citep{DBLP:conf/conll/RatinovR09,DBLP:conf/semweb/Metke-JimenezK16,DBLP:conf/acl/StrakovaSH19,DBLP:conf/acl/DaiKHP20}. While the span-based models need to enumerate all possible spans, which is quadratic to the length of the sentence and is almost impossible to enumerate in the discontinuous NER scenario \citep{DBLP:conf/acl/YuBP20}. Therefore, span-based methods usually will set a maximum span length \citep{DBLP:conf/acl/XuJW17,DBLP:conf/naacl/LuanWHSOH19,DBLP:conf/emnlp/WangL18}. Although hypergraphs can efficiently represent all spans \citep{DBLP:conf/emnlp/LuR15,DBLP:conf/naacl/KatiyarC18,DBLP:conf/emnlp/MuisL16}, it suffers from the spurious structure problem, and structural ambiguity issue during inference and the decoding is quite complicated \citep{DBLP:conf/emnlp/MuisL17}. Because the problems lie in different formulations, no publication has tested their model or framework in three NER subtasks simultaneously to the best of our knowledge.

% 创新思路是什么
In this paper, we propose using a novel and simple sequence-to-sequence (Seq2Seq) framework with the pointer mechanism  \citep{DBLP:conf/nips/VinyalsFJ15} to generate the entity sequence directly. On the source side, the model inputs the sentence, and on the target side, the model generates the entity pointer index sequence. Since flat, continuous and discontinuous entities can all be represented as entity pointer index sequences, this formulation can tackle all the three kinds of NER subtasks in a unified way. Besides, this formulation can even solve the crossing structure entity\footnote{Namely, for span ABCD, both ABC and BCD are entities. Although this is rare, it exists \citep{DBLP:conf/acl/DaiKHP20}.} and multi-type entity\footnote{An entity can have multiple entity types, as proteins can be annotated as drug/compound in the EPPI corpus \citep{DBLP:conf/bionlp/AlexHG07}.}. By converting the NER task into a Seq2Seq generation task, we can smoothly use the Seq2Seq pre-training model BART \citep{DBLP:conf/acl/LewisLGGMLSZ20} to enhance our model.
To better utilize the pre-trained BART, we propose three kinds of entity representations to linearize entities into entity pointer index sequences.

% 结论
Our contribution can be summarized as follows:
\begin{itemize}
\setlength{\itemsep}{1pt}%
\setlength{\parskip}{1pt}%
  \item We propose a novel and simple generative solution to solve the flat NER, nested NER, and discontinuous NER subtasks in a unified framework, in which NER subtasks are formulated as an entity span sequence generation problem.
  \item We incorporate the pre-trained Seq2Seq model BART into our framework and exploit three kinds of entity representations to linearize entities into sequences. The results can shed some light on further exploration of BART into the entity sequence generation.
  \item The proposed framework not only avoids the sophisticated design of tagging schema or span enumeration but also achieves SoTA or near SoTA performance on eight popular datasets, including two flat NER datasets, three nested NER datasets, and three discontinuous NER datasets.
\end{itemize}

\section{Background}

\subsection{NER Subtasks}
% 首先介绍一下NER这个任务是什么. 然后介绍一下nested NER的数据集（genia，ACE04），discountious的数据集
The term ``Named Entity'' was coined in the Sixth Message Understanding Conference (MUC-6) \citep{DBLP:conf/coling/GrishmanS96}. After that, the release of CoNLL-2003 NER dataset has greatly advanced the flat NER subtask \citep{DBLP:conf/conll/SangM03}. \citet{DBLP:conf/ismb/KimOTT03} found that in the field of molecular biology domain, some entities could be nested. \citet{DBLP:journals/jbi/KarimiMKW15} provided a corpus that contained medical forum posts on patient-reported Adverse Drug Events (ADEs), some entities recognized in this corpus may be discontinuous. Despite the difference between the three kinds of NER subtasks, the methods adopted by previous publications can be roughly divided into three types.

\textbf{Token-level classification} The first line of work views the NER task as a token-level classification task, which assigns to each token a tag that usually comes from the Cartesian product between entity labels and the tag scheme, such as BIO and BILOU \citep{DBLP:conf/conll/RatinovR09,DBLP:journals/jmlr/CollobertWBKKK11,DBLP:journals/corr/HuangXY15,DBLP:journals/tacl/ChiuN16,DBLP:conf/naacl/LampleBSKD16,DBLP:conf/bionlp/AlexHG07,DBLP:conf/acl/StrakovaSH19,DBLP:conf/semweb/Metke-JimenezK16,DBLP:conf/emnlp/MuisL17,DBLP:conf/acl/DaiKHP20}, then Conditional Random Fields (CRF) \citep{lafferty2001conditional} or tag sequence generation methods can be used for decoding. Though the work of \citep{DBLP:conf/acl/StrakovaSH19,DBLP:conf/pakdd/WangLZXL19,DBLP:conf/ijcai/ZhangCZLY18,DBLP:conf/coling/ChenM18} are much like our method, they all tried to predict a tagging sequence. Therefore, they still need to design tagging schemas for different NER subtasks.

\textbf{Span-level classification} When applying the sequence labelling method to the nested NER and discontinous NER subtasks, the tagging will be complex \citep{DBLP:conf/acl/StrakovaSH19,DBLP:conf/semweb/Metke-JimenezK16} or multi-level \citep{DBLP:conf/naacl/JuMA18,DBLP:conf/acl/FisherV19,DBLP:journals/tacl/ShibuyaH20}. Therefore, the second line of work directly conducted the span-level classification. The main difference between publications in this line of work is how to get the spans. \citet{DBLP:conf/emnlp/FinkelM09} regarded the parsing nodes as a span. \citet{DBLP:conf/acl/XuJW17,DBLP:conf/naacl/LuanWHSOH19,DBLP:conf/emnlp/YamadaASTM20,DBLP:conf/acl/LiFMHWL20,DBLP:conf/acl/YuBP20,DBLP:conf/acl/WangSCC20} tried to enumerate all spans. Following \citet{DBLP:conf/emnlp/LuR15}, hypergraph methods which can effectively represent exponentially many possible nested mentions in a sentence have been extensively studied in the NER tasks \citep{DBLP:conf/naacl/KatiyarC18,DBLP:conf/emnlp/WangL18,DBLP:conf/emnlp/MuisL16}.

\textbf{Combined token-level and span-level classification} To avoid enumerating all possible spans and incorporate the entity boundary information into the model, \citet{DBLP:conf/emnlp/WangL19,DBLP:conf/emnlp/ZhengCXLX19,DBLP:conf/acl/LinLHS19,DBLP:conf/emnlp/WangLTZ20,DBLP:conf/acl/LuoZ20} proposed combining the token-level classification and span-level classification.

% Unlike previous work, we tackle different NER tasks as a generation task. Our model neither needs to assign labels to each token, which can avoid the design of tagging schema for each NER tasks, nor needs to enumerate spans usually cause high computation overhead.

\subsection{Sequence-to-Sequence Models}
% 这里参考一下BART/T5等的写法
The Seq2Seq framework has been long studied and adopted in NLP \citep{DBLP:conf/nips/SutskeverVL14,DBLP:conf/emnlp/ChoMGBBSB14,DBLP:conf/emnlp/LuongPM15,vaswani2017attention,DBLP:conf/nips/VinyalsFJ15}. \citet{DBLP:conf/naacl/GillickBVS16} proposed a Seq2Seq model to predict the entity's start, span length and label for the NER task. Recently, the amazing performance gain achieved by PTMs (pre-trained models) \citep{DBLP:journals/corr/abs-2003-08271,DBLP:conf/naacl/PetersNIGCLZ18,DBLP:conf/naacl/DevlinCLT19,DBLP:journals/corr/abs-2104-04986,DBLP:journals/tacl/YanQH20} has attracted several attempts to pre-train a Seq2Seq model \citep{DBLP:conf/icml/SongTQLL19,DBLP:conf/acl/LewisLGGMLSZ20,DBLP:journals/jmlr/RaffelSRLNMZLL20}. We mainly focus on the newly proposed BART \citep{DBLP:conf/acl/LewisLGGMLSZ20} model because it can achieve better performance than MASS \citep{DBLP:conf/icml/SongTQLL19}. And the sentence-piece tokenization used in T5 \citep{DBLP:journals/jmlr/RaffelSRLNMZLL20} will cause different tokenizations for the same token, making it hard to generate pointer indexes to conduct the entity extraction. 

BART is formed by several transformer encoder and decoder layers, like the transformer model used in the machine translation \citep{vaswani2017attention}. BART's pre-training task is to recover corrupted text into the original text. BART uses the encoder to input the corrupted sentence and the decoder to recover the original sentence. BART has base and large versions. The base version has 6 encoder layers and 6 decoder layers, while the large version has 12. Therefore, the number of parameters is similar to its equivalently sized BERT \footnote{Because of the cross-attention between encoder and decoder, the number of parameters of BART is about 10\% larger than its equivalently sized of BERT \citep{DBLP:conf/acl/LewisLGGMLSZ20}.}.
%However, unlike RoBERTa which is trained on the masked language model, BART is trained in an autoregressive way.

\begin{figure*}[t]
  \centering
  \includegraphics[width=\textwidth]{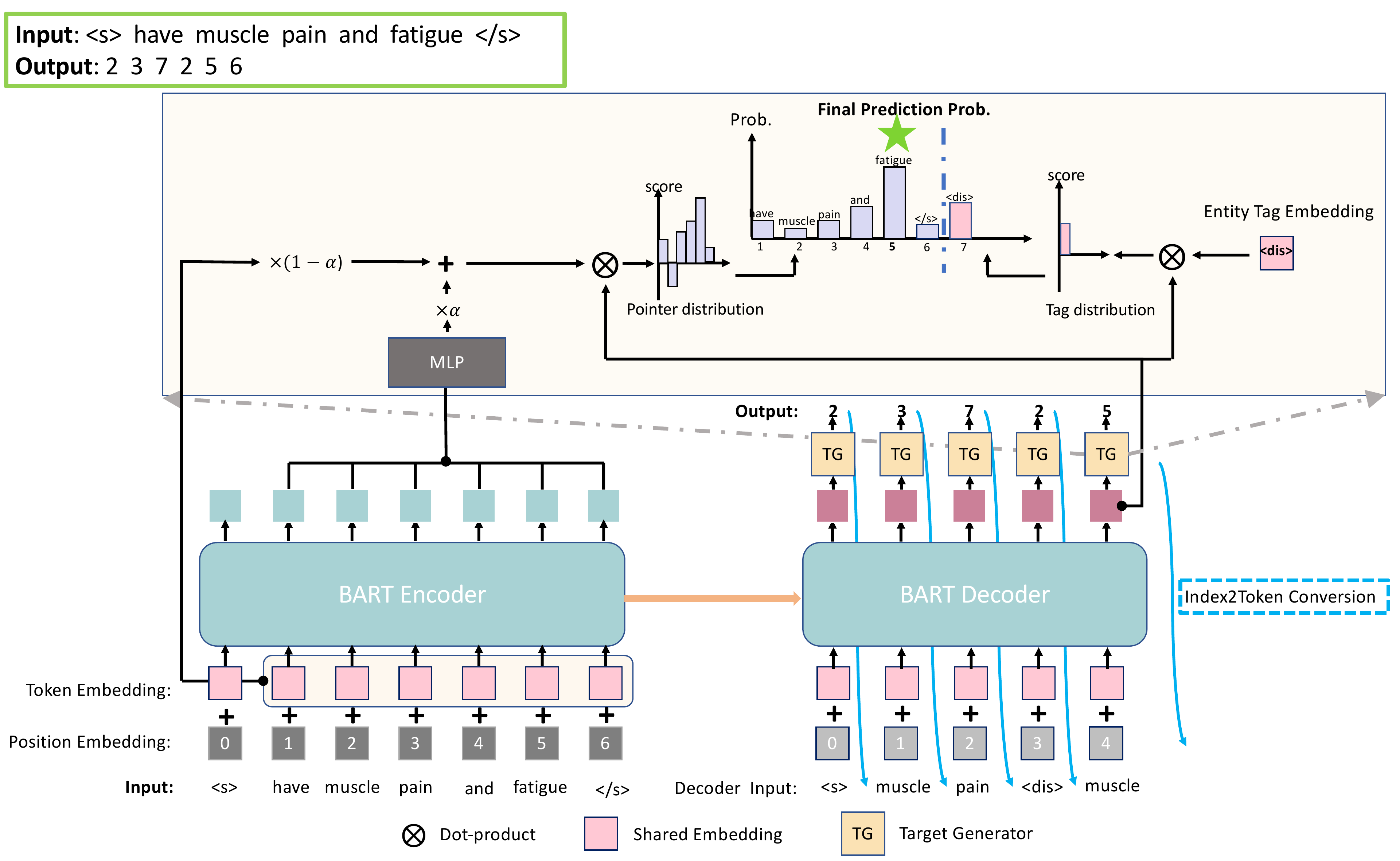}
  \caption{Model structure used in our method. The encoder encodes input sentences, and the decoder uses the pointer mechanism to generate indexes autoregressively. ``$<$s$>$'' and ``$<$/s$>$'' are the predefined start-of-sentence and end-of-sentence tokens in BART. In the output sequence, ``7'' means the entity tag ``$<$dis$>$'', and other numbers indicate the pointer index (in range [1, 6]).}\label{fig:strucutre}
\end{figure*}

\section{Proposed Method}
% 一上来直接介绍一下，在我们这里是如何formulate NER的
In this part, we first introduce the task formulation, then we describe how we use the Seq2Seq model with the pointer mechanism to generate the entity index sequences. After that, we present the detailed formulation of our model with BART.
% we present the entity representations we used to linearize entities into ordered sequence and how to use BART to generate this ordered target sequence.

\subsection{NER Task}
The three kinds of NER tasks can all be formulated as follows, given an input sentence of $n$ tokens $X=[x_1, x_2, ..., x_n]$,  the target sequence is $Y=[s_{11}, e_{11},...,s_{1j}, e_{1j}, t_1, ..., s_{i1}, e_{i1},...,s_{ik},e_{ik}, t_i]$, where $s,e$ are the start and end index of a span, since an entity may contain one (for flat and nested NER) or more than one (for discontinuous NER) spans, each entity is represented as $[s_{i1}, e_{i1},...,s_{ij}, e_{ij}, t_i]$, where $t_i$ is the entity tag index. We use $G=[g_1, ..., g_l]$ to denote the entity tag tokens (such as ``Person'', ``Location'', etc.), where $l$ is the number of entity tags.  We make $t_i \in (n, n+l]$, the $n$ shift is to make sure $t_i$ is not confusing with pointer indexes (pointer indexes will be in range $[1, n]$).

\subsection{Seq2Seq for Unified Decoding}
Since we formulate the NER task in a generative way, we can view the NER task as the following equation:
\begin{align}
  P(Y|X) = \prod_{t=1}^m P(y_t|X, Y_{<t})
\end{align}
where $y_0$ is the special ``start of sentence'' control token.

We use the Seq2Seq framework with the pointer mechanism to tackle this task. Therefore, our model consists of two components:

(1) \textbf{Encoder} encodes the input sentence $X$ into vectors $\mathbf{H}^e$, which formulates as follows:
\begin{align}
  \mathbf{H}^e = \mathrm{Encoder}(X)
\end{align}
where $\mathbf{H}^e \in \mathbb{R}^{n \times d}$, and $d$ is the hidden dimension.

(2) \textbf{Decoder} is to get the index probability distribution for each step $P_t = P(y_t|X, Y_{<t})$. However, since $Y_{<t}$ contains the pointer and tag index, it cannot be directly inputted to the Decoder. We use the Index2Token conversion to convert indexes into tokens
\begin{align}
 \hat{y}_t = \begin{cases}
    X_{y_t},&  \text{if}  \  y_t \le n,\\
    G_{y_t-n},&  \text{if} \  y_t > n
  \end{cases}
\end{align}

After converting each $y_t$ this way, we can get the last hidden state $\mathbf{h}_t^d \in \mathbb{R}^d$ with $\hat{Y}_{<t}=[\hat{y}_1, ..., \hat{y}_{t-1}]$ as follows
\begin{align}
  \mathbf{h}_t^d = \mathrm{Decoder}(\mathbf{H}^e; \hat{Y}_{<t})
\end{align}
Then, we can use the following equations to achieve the index probability distribution $P_t$
\begin{align}
  \mathbf{E}^e & = \mathrm{TokenEmbed}(X)\\
  \mathbf{\hat{H}}^e & = \mathrm{MLP}(\mathbf{H}^e) \\
  \mathbf{\bar{H}}^e & = \alpha*\mathbf{\hat{H}}^e + (1-\alpha)*\mathbf{E}^e \\
  \mathbf{G}^d & = \mathrm{TokenEmbed}(G) \\
  P_t & = \mathrm{Softmax}([\mathbf{\bar{H}^e}\otimes \mathbf{h}_t^d;\mathbf{G}^d\otimes \mathbf{h}_t^d] )
\end{align}
where TokenEmbed is the embeddings shared between the Encoder and Decoder; $\mathbf{E}^e,\mathbf{\hat{H}}^e,\mathbf{\bar{H}}^e \in \mathbb{R}^{n \times d}$; $\alpha \in \mathbb{R}$ is a hyper-parameter; $\mathbf{G}^d \in \mathbb{R}^{l \times d}$; $[\, \cdot \, ; \, \cdot \, ]$ means concatenation in the first dimension; $\otimes$ means the dot product.

% The reason why we add $\mathbf{E}^e$ and $\mathbf{H}^e$ together to represent the input token is that we want the generation process to be more similar to the pre-training process of BART.

During the training phase, we use the negative log-likelihood loss and the teacher forcing method. During the inference, we use an autoregressive manner to generate the target sequence. We use the decoding algorithm presented in Algorithm \ref{al} to convert the index sequence into entity spans.

\begin{algorithm}[!bht]
  \begin{algorithmic}[1]
    \caption{Decoding Algorithm to Convert the Entity Representation Sequence into Entity Spans} \label{al}
    \Require Target sequence $Y=[y_1, ..., y_{m}]$ and $y_i \in [1, n+|G|]$
    \Ensure  Entity spans $E=\{(e_1,t_1), ..., (e_i,t_i)\}$
    \State $E=\{\}, e=[], i=1$
    \While{$i<=m$}
      \State $y_i = Y[i]$
      \If{$y_i > n$}
        \If{$len(e)>0$}
          \State $E.add((e, G_{y_i-n}))$
        \EndIf
        \State $e=[]$
      \Else
      \State $e.append(y_i)$
      \EndIf
      \State $i=i+1$
    \EndWhile
    \State \Return $E$
  \end{algorithmic}
\end{algorithm}

\begin{figure}[!bht]
  \centering
  \includegraphics[width=\columnwidth]{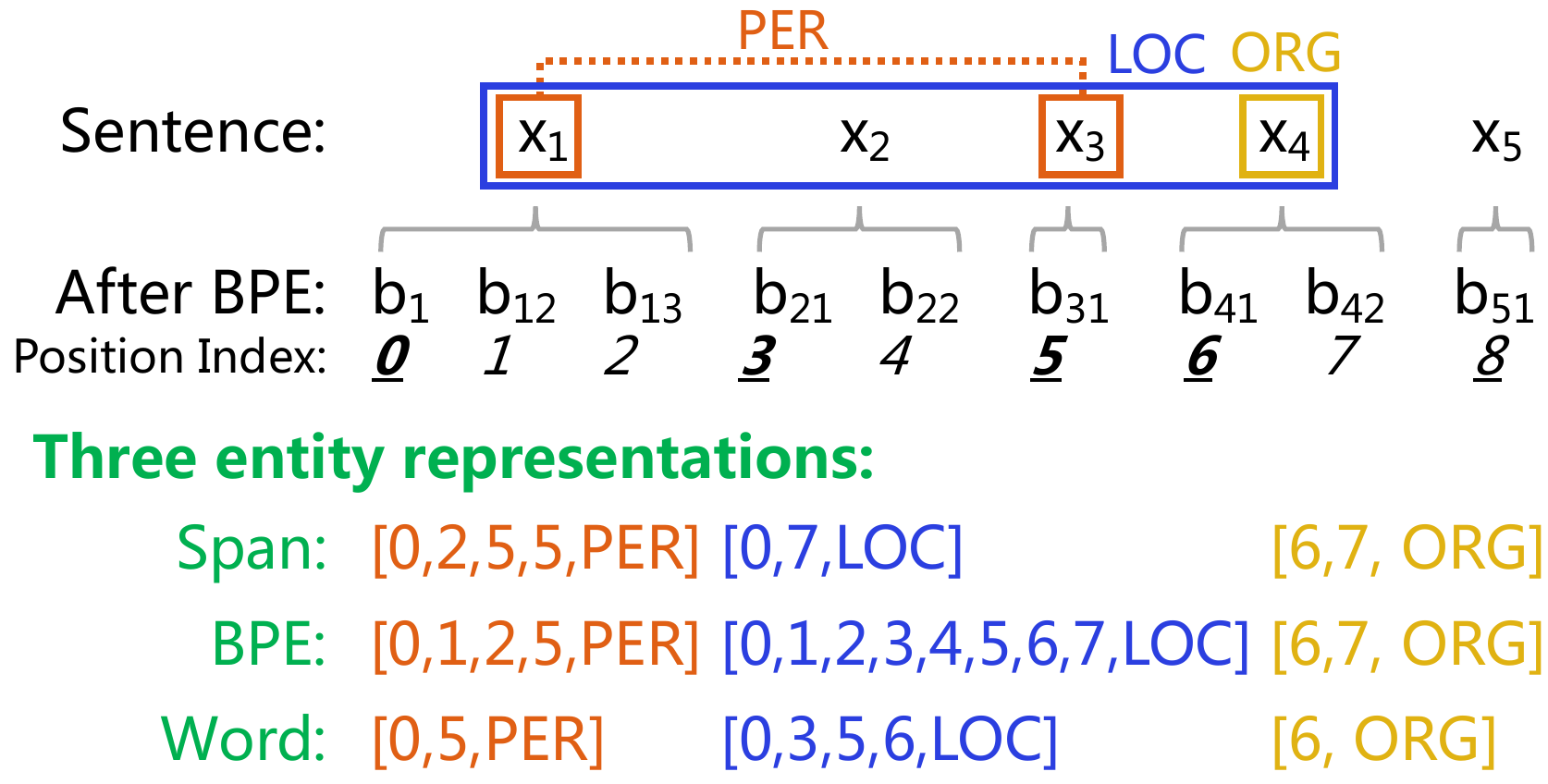}
  \caption{The bottom three lines are examples of the three kinds of entity representations to determine the entity in the sentence unambiguously. Words in the boxes are entity words, words within the same color box belong to the same entity, and their corresponding entity representation is also with the same color. There are three entities, $(x_1, x_3, PER)$, $(x_1, x_2, x_3, x_4, LOC)$, $(x_4, FAC)$, where $LOC,PER,FAC$ are their corresponding entity tags. The underlined position index means this is the starting BPE of a word.}\label{fig:span_representation}
\end{figure}

\subsection{Detailed Entity Representation with BART}
% todo 加入对tokenization方法的描述
Since our model is a Seq2Seq model, it is natural to utilize the pre-training Seq2Seq model BART to enhance our model. We present a visualization of our model based on BART in Figure~\ref{fig:strucutre}. However, BART's adoption is non-trivial because the Byte-Pair-Encoding (BPE) tokenization used in BART might tokenize one token into several BPEs. To exploit how to use BART efficiently, we propose three kinds of pointer-based entity representations to locate entities in the original sentence unambiguously. The three entity representations are as follows:

\textbf{Span} The position index of the first BPE of the starting entity word and the last BPE of the ending entity word. If this entity includes multiple discontinuous spans of words, each span is represented in the same way.

\textbf{BPE} The position indexes of all BPEs of the entity words.

\textbf{Word} Only the position index of the first BPE of each entity word is used.

For all cases, we will append the entity tag to the entity representation. An example of the entity representations is presented in Figure~\ref{fig:span_representation}. If a word does not belong to any entity, it will not appear in the target sequence. If a whole sentence has no entity, the prediction should be an empty sequence (only contains the ``start of sentence'' ($<$s$>$) token and the ``end of sentence'' ($<$/s$>$) token ).

\section{Experiment}

\subsection{Datasets}
To show that our proposed method can be used in various NER subtasks, we conducted experiments on eight datasets.

\textbf{Flat NER Datasets} We adopt the CoNLL-2003 \citep{DBLP:conf/conll/SangM03} and the OntoNotes dataset \footnote{\url{https://catalog.ldc.upenn.edu/LDC2013T19}} \citep{DBLP:conf/conll/PradhanMXNBUZZ13}. For CoNLL-2003, we follow \citet{DBLP:conf/naacl/LampleBSKD16,DBLP:conf/acl/YuBP20} to train our model on the concatenation of the train and development sets. For the OntoNotes dataset, we use the same train, development, test splits as \citet{DBLP:conf/conll/PradhanMXUZ12,DBLP:conf/acl/YuBP20}, and the New Testaments portion were excluded since there is no entity in this portion \citep{DBLP:journals/tacl/ChiuN16}.

\textbf{Nested NER Datasets} We conduct experiments on ACE 2004\footnote{\url{https://catalog.ldc.upenn.edu/LDC2005T09}} \citep{DBLP:conf/lrec/DoddingtonMPRSW04}, ACE 2005\footnote{\url{https://catalog.ldc.upenn.edu/LDC2006T06}} \citep{walker2005ace}, Genia corpus \citep{DBLP:conf/ismb/KimOTT03}. For ACE2004 and ACE2005, we use the same data split as \citet{DBLP:conf/emnlp/LuR15,DBLP:conf/emnlp/MuisL17,DBLP:conf/acl/YuBP20}, the ratio between train, development and test set is 8:1:1. For Genia, we follow \citet{DBLP:conf/emnlp/WangLTZ20,DBLP:journals/tacl/ShibuyaH20} to use five types of entities and split the train/dev/test as 8.1:0.9:1.0.

\begin{table*}[ht]
  \centering
  \setlength{\tabcolsep}{2pt}
  \begin{tabular}{c|ccc|ccc}
    % \Xhline{1.2pt}
    \toprule
     & \multicolumn{3}{c|}{CoNLL2003} & \multicolumn{3}{c}{OntoNotes} \\
  Models  & P        & R        & F       & P        & R        & F       \\
    % \Xhline{0.8pt}
    \midrule
  \citet{DBLP:conf/emnlp/ClarkLML18}[GloVe300d]                      & -        & -        & 92.6    &    -     &      -   &     -    \\
  \citet{DBLP:conf/naacl/PetersNIGCLZ18}[ELMo] & - & - & 92.22 & - & - & - \\
  \citet{DBLP:conf/naacl/AkbikBV19}[Flair] & -  & - & 93.18 & - & - & - \\
  \citet{DBLP:conf/acl/StrakovaSH19}[BERT-Large] & -        & -        & 93.07   &      -   &       -  &    -     \\
  \citet{DBLP:conf/emnlp/YamadaASTM20}[RoBERTa-Large] & -        & -        & 92.40   &      -   &       -  &    -     \\
  \citet{DBLP:conf/acl/LiFMHWL20}[BERT-Large]$\dagger$ &  92.47 & 93.27 & 92.87 &  \textbf{91.34} & 88.39 & 89.84 \\
  \citet{DBLP:conf/acl/YuBP20}[BERT-Large]$\ddagger$            & \textbf{92.85}    & 92.15    & 92.5    & 89.92    & 89.74    & 89.83   \\
  \midrule
  Ours(Span)[BART-Large]    & 92.31    & 93.45    & 92.88   & 88.94    & 90.33    & 89.63   \\
  Ours(BPE)[BART-Large]  & 92.60    & 93.22    & 92.96   & 90.00    & 89.52    & 89.76   \\
  Ours(Word)[BART-Large]    & 92.61    & \textbf{93.87}    & \textbf{93.24}   & 89.99    & \textbf{90.77}    & \textbf{90.38}   \\
  \bottomrule
  \end{tabular}
  \caption{Results for the flat NER datasets. ``$\dagger$'' indicates we rerun their code. ``$\ddagger$'' means our reproduction with only the sentence-level context \protect\footnotemark. }
  \label{tb:flat_ner}
\end{table*}
\footnotetext{In the reported experiments, they included the document context. We rerun their code with only the sentence context. The lack of document context might cause performance degradation is also confirmed by the author himself in \url{https://github.com/juntaoy/biaffine-ner/issues/8\#issuecomment-650813813}. \label{fn:yu}}

\begin{table*}[ht]
  \centering
  \setlength{\tabcolsep}{2pt}
  \begin{tabular}{c|ccc|ccc|ccc}
    % \Xhline{1.2pt}
    \toprule
    & \multicolumn{3}{c|}{ACE2004} & \multicolumn{3}{c|}{ACE2005} & \multicolumn{3}{c}{Genia} \\
   Models & P       & R       & F       & P       & R       & F       & P       & R      & F      \\
    % \Xhline{0.8pt}
    \midrule
  \citet{DBLP:conf/naacl/LuanWHSOH19}[ELMO]       & -       & -       & 84.7    & -       & -       & 82.9    & -       & -      & 76.2   \\
  \citet{DBLP:conf/acl/StrakovaSH19}[BERT-Large] & -       & -       & 84.33   & -       & -       & 83.42   & -       & -      & 76.44  \\
  \citet{DBLP:journals/tacl/ShibuyaH20}[BERT-Large]$\star$ & 85.23 &  84.72 & 84.97      & 83.30   & 84.69   & 83.99   & 77.46   & 76.65  & 77.05  \\
  \citet{DBLP:conf/acl/LiFMHWL20}[BERT-Large]$\dagger$ & 85.83   & 85.77   &  85.80  & \textbf{85.01}   & 84.13   & 84.57   & \textbf{81.25}   & 76.36  &  78.72 \\
  \citet{DBLP:conf/acl/YuBP20}[BERT-Large] $\ddagger$  & 85.42   & 85.92   & 85.67   & 84.50   & 84.72   & 84.61   & 79.43   & 78.32  & 78.87  \\
  \citet{DBLP:conf/acl/WangSCC20}[BERT-Large]$\star$     & 86.08   & \textbf{86.48}   & 86.28   & 83.95   & 85.39   & 84.66   & 79.45   & 78.94  & 79.19  \\
  % \hdashline
  \midrule
  Ours(Span)[BART-Large]                          & 84.81   & 83.64   & 84.22   & 81.41   & 83.24   & 82.31   & 78.87   & \textbf{79.6}   & \textbf{79.23}  \\
  Ours(BPE)[BART-Large] & 86.69   & 83.83   & 85.24   & 82.08   & 83.44   & 82.75   & 78.15   & 79.06  & 78.60  \\
  Ours(Word)[BART-Large] & \textbf{87.27}   & 86.41   & \textbf{86.84}   & 83.16   & \textbf{86.38}   & \textbf{84.74}   & 78.57   & 79.3   & 78.93  \\
  % \Xhline{1.2pt}
  \bottomrule
  \end{tabular}
  \caption{Results for nested NER datasets,``$\dagger$'' means our rerun of their code. ``$\ddagger$'' means our reproduction with only sentence-level context\footref{fn:yu}. ``$\star$'' for a fair comparison, we only present results with the BERT-Large model.}
  \label{tb:nested_ner}
\end{table*}

\textbf{Discontinuous NER Datasets} We follow \citet{DBLP:conf/acl/DaiKHP20} to use CADEC \citep{DBLP:journals/jbi/KarimiMKW15}, ShARe13 \citep{DBLP:conf/clef/PradhanESMCVSCS13} and ShARe14 \citep{DBLP:conf/clef/MoweryVSCMKGEPSC14} corpus. Since only the Adverse Drug Events (ADEs) entities include discontinuous annotation, only these entities were considered \citep{DBLP:conf/acl/DaiKHP20,DBLP:conf/semweb/Metke-JimenezK16,DBLP:journals/wicomm/TangHWC18}.

\subsection{Experiment Setup}
% 说一下优化器，学习率
We use the BART-Large model, whose encoder and decoder each has 12 layers for all experiments, making it the same number of transformer layers as the BERT-Large and RoBERTa-Large model. We did not use any other embeddings, and the BART model is fine-tuned during the optimization.  We put more detailed experimental settings in the Supplementary Material. We report the span-level F1.

\begin{table*}[!htb]
  \centering
  \setlength{\tabcolsep}{2pt}  % 设置column艰巨
  \begin{tabular}{c|ccc|ccc|ccc}
    % \Xhline{1.2pt}
    \toprule
  & \multicolumn{3}{c|}{CADEC} & \multicolumn{3}{c|}{ShARe13} & \multicolumn{3}{c}{ShARe14} \\
  Model & P       & R      & F      & P        & R       & F       & P        & R       & F       \\
  % \Xhline{0.8pt}
  \midrule
  \citet{DBLP:conf/semweb/Metke-JimenezK16} & 64.4    & 56.5   & 60.2   & -        & -       & -       & -        & -       & -       \\
  \citet{DBLP:journals/wicomm/TangHWC18}    & 67.8    & 64.9   & 66.3   & -        & -       & -       & -        & -       & -       \\
  \citet{DBLP:conf/acl/DaiKHP20}[ELMo]     & 68.9    & 69.0   & 69.0   & 80.5     & 75.0    & 77.7    & \textbf{78.1}     & 81.2    & 79.6    \\
  % \hdashline
  \midrule
  Ours(Span)[BART-Large]    & \textbf{71.55}  & 68.59 & 70.04  & 80.42    & \textbf{78.15}   & 79.27   & 76.85    & 83.59   & 80.08   \\
  Ours(BPE)[BART-Large]  & 69.45   & 70.51  & 69.97  & 82.07    & 76.45   & 79.16   & 75.88    & \textbf{84.37}   & 79.90   \\
  Ours(Word)[BART-Large] & 70.08   & \textbf{71.21}  & \textbf{70.64}  & \textbf{82.09}    & 77.42   & \textbf{79.69}   & 77.2     & 83.75   & \textbf{80.34}  \\
  % \Xhline{1.2pt}
  \bottomrule
  \end{tabular}
  \caption{Results for discontinuous NER datasets. }
  \label{tb:dis_ner}
\end{table*}

\section{Results}
\subsection{Results on Flat NER}
Results are shown in Table \ref{tb:flat_ner}. We do not compare with \citet{DBLP:conf/emnlp/YamadaASTM20} since they added entity information during the pre-training process. \citet{DBLP:conf/emnlp/ClarkLML18,DBLP:conf/naacl/PetersNIGCLZ18,DBLP:conf/naacl/AkbikBV19,DBLP:conf/acl/StrakovaSH19} assigned a label to each token, and \citet{DBLP:conf/acl/LiFMHWL20,DBLP:conf/acl/YuBP20} are based on span-level classifications, while our method is based on the entity sequence generation. And for both datasets, our method achieves better performance. We will discuss the performance difference between our three entity representations in Section \ref{sec:three_spec}.

\begin{table*}[bth]
  \centering
  \setlength{\tabcolsep}{2pt}  % 设置column间距
  \renewcommand{\arraystretch}{1.2}
  \begin{tabular}{c|c|c|c|c|c|c|c|c}
    % \Xhline{0.08em}
    % \Xhline{1.2pt}
    \toprule
     Entity& \multicolumn{2}{c|}{Flat NER} & \multicolumn{3}{c|}{Nested NER} & \multicolumn{3}{c}{Discontinuous NER} \\
     \cline{2-9}
    % \cmidrule(r){2-9}
   Representation & CoNLL2003 & OntoNotes & ACE2004  & ACE2005  & Genia    & CADEC    & ShARe13 & ShARe14 \\
  %  \Xhline{0.8pt}
  \midrule
   Span & 3.0/3.0   & 3.0/3.0   & \textbf{3.0/3.0}  & \textbf{3.0/3.0}  & \textbf{3.0/3.0}  & 3.17/3.0 & 3.15/3.0  & \textbf{3.2/3.0} \\
  BPE  & 3.55/3.0  & 3.39/3.0  & 4.15/3.0 & 3.84/3.0 & 5.21/5.0 & 4.08/4.0 & 3.92/3.0 & 4.34/4.0 \\
  Word & \textbf{2.44/2.0}  & \textbf{2.86/2.0}  & 3.53/2.0  & 3.26/2.0 & 3.09/3.0 & \textbf{2.72/3.0} & \textbf{2.63/3.0} & 3.74/3.0 \\
  % \Xhline{1.2pt}
  \bottomrule
  \end{tabular}
  \caption{The average (before /) and median entity length (including the entity label) for each entity representations in the respective testing set.}
  \label{tb:avg_length}
\end{table*}

\subsection{Results on Nested NER}
Table \ref{tb:nested_ner} presents the results for the three nested NER datasets, and our proposed BART-based generative models are comparable to the token-level classication \citep{DBLP:conf/acl/StrakovaSH19,DBLP:journals/tacl/ShibuyaH20} and span-level classification \citep{DBLP:conf/naacl/LuanWHSOH19,DBLP:conf/acl/LiFMHWL20,DBLP:conf/acl/WangSCC20} models.

\subsection{Results on Discontinuous NER}
Results in Table \ref{tb:dis_ner} show the comparison between our model and other models in three discontinuous NER datasets. Although \citet{DBLP:conf/acl/DaiKHP20} tried to utilize BERT to enhance the model performance, they found that ELMo worked better. In all three datasets, our model achieves better performance.

\subsection{Comparison Between Different Entity Representations} \label{sec:three_spec}
In this part, we discuss the performance difference between the three entity representations.
The ``Word'' entity representation achieves better performance almost in all datasets. And the comparison between the ``Span'' and ``BPE'' representations is more involved. To investigate the reason behind these results, we calculate the average and median length of entities when using different entity representations, and the results are presented in Table \ref{tb:avg_length}. It is clear that for a generative framework, the shorter the entity representation the better performance it should achieve. Therefore, as shown in Table \ref{tb:avg_length}, the ``Word'' representation with smaller average entity length in CoNLL2003, OntoNotes, CADEC, ShARe13 achieves better performance in these datasets. However, although the average entity length of the ``BPE'' representation is longer than the ``Span'' representation, it achieves better performance in CoNLL2003, OntoNotes, ACE2004, ACE2005, this is because the ``BPE'' representation is more similar to the pre-training task, namely, predicting continuous BPEs. And we believe this task similarity is also the reason why the ``Word'' representation (Most of the words will be tokenized into a single BPE, making the ``Word'' representation still continuous.) achieves better performance than the ``Span'' representation in ACE2004, ACE2005, and ShARe14, although the former has longer entity length.

\begin{table*}[!bht]
  \centering
  \setlength{\tabcolsep}{2pt}  % 设置column间距
  \renewcommand{\arraystretch}{1.2}
  \begin{tabular}{c|c|c|c|c|c|c|c|c}
    % \Xhline{1.2pt}
    \toprule
         & \multicolumn{2}{c|}{Flat NER} & \multicolumn{3}{c|}{Nested NER} & \multicolumn{3}{c}{Discontinuous NER} \\
    \cline{2-9}
  Errors & CoNLL2003     & OntoNotes    & ACE2004   & ACE2005   & Genia  & CADEC     & ShARe13     & ShARe14     \\
  % \Xhline{0.8pt}
  \hline
  $E_1$     & 0.05\%        & 0.02\%       & 0.23\%    & 0.06\%   & 0.0\%   & 0.31\%     & 0.0\%       & 0.01\%     \\  \hline

  $E_2$     & 0.04\%        & 0.03\%       & 0.13\%    & 0.22\%   & 0.11\%  & 1.02\%     & 0.18\%      & 0.16\%     \\  \hline
  $E_3$     & 0.05\%        & 0.02\%       & 0.30\%    & 0.26\%   & 0.06\%  & 0.0\%      & 0.08\%      & 0.02\%   \\
  % \Xhline{1.2pt}
  \bottomrule
  \end{tabular}
  \caption{Different invalid prediction probability for the ``Word'' entity representation. $E_1$ means the predicted indexes contain index which is not the start index of a word, $E_2$ means the predicted indexes within an entity are not increasing, $E_3$ means duplicated entity prediction.  }
  \label{tb:invalid}
\end{table*}

\begin{figure*}[!t]
  \centering
  \begin{subfigure}{0.32\textwidth}
      \begin{tikzpicture}[scale=0.6]
      % \draw[very thick,blue] (3,-4) ellipse ({4} and {3});

      \begin{axis}[
          xlabel={Entity Position(\# of Entities)},
          y label style={at={(axis description cs:0.05, 0.5)}},
          ylabel={Recall},
          height=0.9\textwidth,
          width=1.7\textwidth,
          xtick={1, 2, 3, 4, 5, 6, 7},
          xticklabels={1(4.6),2(2.7),3(1.7),4(1),5(0.6),6+(0.7)},
          grid=major,
          legend pos=south east,
          title=OntoNotes
      ]

      \addplot+[color=blue,mark=triangle] coordinates {
      (1, 90.2)  (2, 90.89)  (3, 90.94)
      (4, 90.94)  (5, 92.01)
      (6, 92.84)
      };
      \end{axis}

      \end{tikzpicture}
  \end{subfigure}
  % \hspace{-0.5cm}
  \begin{subfigure}{0.32\textwidth}
      \begin{tikzpicture}[scale=0.6]
      % \draw[very thick,blue] (3,-4) ellipse ({4} and {3});

      \begin{axis}[
          xlabel={Entity Position(\# of Entities)},
          y label style={at={(axis description cs:0.05, 0.5)}},
          ylabel={Recall},
          xtick={1, 2, 3, 4, 5, 6, 7},
          xticklabels={1(1.6),2(1.3),3(1.0),4(0.7),5(0.4),6+(0.5)},
          height=0.9\textwidth,
          width=1.7\textwidth,
          grid=major,
          legend pos=south east,
          title=Genia,
      ]

      \addplot+[color=blue,mark=triangle] coordinates {
      (1, 78.43)  (2, 79.85)  (3, 74.72)
      (4, 71.78)  (5, 77.86)
      (6, 77.57)
      };
      \end{axis}
      \end{tikzpicture}
  \end{subfigure}
  \begin{subfigure}{0.32\textwidth}
      \begin{tikzpicture}[scale=0.6]
      % \draw[very thick,blue] (3,-4) ellipse ({4} and {3});

      \begin{axis}[
          xlabel={Entity Position(\# of Entities)},
          y label style={at={(axis description cs:0.05, 0.5)}},
          ylabel={Recall},
          xtick={1, 2, 3, 4, 5, 6, 7},
          xticklabels={1(4.9),2(1.7),3(0.7),4(0.3),5(0.1),6+(0.2)},
          height=0.9\textwidth,
          width=1.7\textwidth,
          grid=major,
          legend pos=south east,
          title=ShARe14
      ]

      \addplot+[color=blue,mark=triangle] coordinates {
      (1, 84.05)  (2, 82.81)  (3, 84.03)
      (4, 85.02)  (5, 89.54)
      (6, 91.38)
      };
      \end{axis}

      \end{tikzpicture}
  \end{subfigure}
  \caption{The recall of entities in different entity sequence positions, the number of entities in that position is the number in the bracket (the unit is 1000). } \label{fig:entity_order}
\end{figure*}
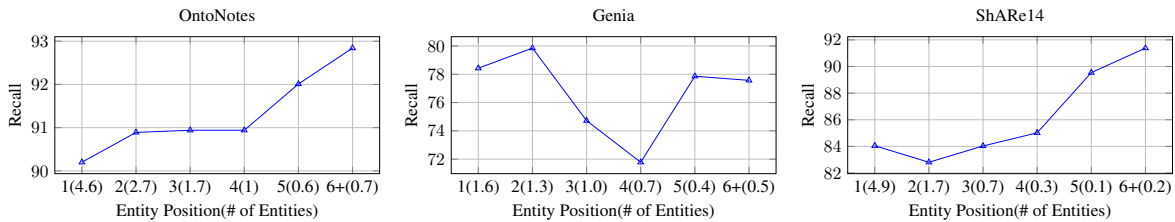

A clear outlier is the Genia dataset, where the ``Span'' representation achieves better performance than the other two. We presume this is because in this dataset, a word will be tokenized into a longer BPE sequence (this can be inferred from the large entity length gap between the ``Word'' and ``BPE'' representation.) so that the ``Word'' representation will also be dissimilar to the pre-training tasks. For example, the protein ``lipoxygenase isoforms'' will be tokenized into the sequence ``[`Ġlip', `oxy', `gen', `ase', `Ġiso', `forms']'', which makes the target sequence of the ``Word'' representation be ``[`Ġlip', `Ġiso']'', resulting a discontiguous BPE sequence. Therefore, the shorter ``Span'' representation achieves better performance in this dataset.

\section{Analysis}
% flat NER选择ontonotes, nested选择ACE2004, Discontinuous选择ShARe13吧

\subsection{Recall of Discontinuous Entities}
Since only about 10\% of entities in the discontinuous NER datasets are discontinuous, only evaluating the whole dataset may not show our model can recognize the discontinuous entities. Therefore, like in \citet{DBLP:conf/acl/DaiKHP20,DBLP:conf/emnlp/MuisL16} we report our model's performance on the discontinuous entities in Table \ref{tb:only_dis}. As shown in Table \ref{tb:only_dis}, our model can predict the discontinuous named entities and achieve better performance.

{\tiny{\begin{table}[!hb]
  \centering
  \setlength{\tabcolsep}{2pt}  % 设置column间距
  \begin{tabular}{c|ccc|ccc}
    % \Xhline{1.2pt}
    \toprule
      & \multicolumn{3}{c|}{ShARe13} & \multicolumn{3}{c}{ShARe14} \\
  Model    & P       & R       & F       & P       & R       & F       \\
  % \Xhline{0.8pt}
  \midrule
  \citet{DBLP:conf/acl/DaiKHP20} & \textbf{78.5}    & 39.4    & 52.5    & \textbf{56.1}    & 43.8    & 49.2    \\
  Ours(Word)  & 57.5   & \textbf{52.8}   & \textbf{55.0}   & 49.6   & \textbf{56.2}   & \textbf{52.7}   \\
  % \Xhline{1.2pt}
  \bottomrule
  \end{tabular}
  \caption{Performance on the discontinuous entities of the tesing dataset of ShARe13 and ShARe14.}
  \label{tb:only_dis}
  \end{table}
}}

\subsection{Invalid Prediction}
In this part, we mainly focus on the analysis of the ``Word'' representation since it generally achieves better performance. We do not restrict the output distribution; therefore, the entity prediction may contain invalid predictions as show in Table \ref{tb:invalid}, this table shows that the BART model can learn the prediction representations quite well since, in most cases, the invalid prediction is less than 1\%. We exclude all these invalid predictions during evaluation.

\subsection{Entity Order Vs. Entity Recall}
Its appearance order in the sentence determines the entity order, and we want to study whether the entity that appears later in the target sequence will have worse recall than entities that appear early. The results are provided in Figure~\ref{fig:entity_order}. The latter the entity appears, the larger probability that it can be recalled for the flat NER and discontinuous NER. While for the nested NER, the recall curve is quite involved. We assume this phenomenon is because, for the flat NER and discontinuous NER (more than 91.1\% of entities are continuous) datasets, different entities have less dependence on each other. While in the nested NER dataset, entities in the latter position may be the outermost entity that contains the former entities. The wrong prediction of former entities may negatively influence the later entities.

\section{Conclusion}
In this paper, we formulate NER subtasks as an entity span sequence generation problem, so that we can use a unified Seq2Seq model with the pointer mechanism to tackle flat, nested, and discontinuous NER subtasks. The Seq2Seq formulation enables us to smoothly incorporate the pre-training Seq2Seq model BART to enhance the performance. To better utilize BART, we test three types of entity representation methods to linearize the entity span into sequences. Results show that the entity representation with a shorter length and more similar to continuous BPE sequences achieves better performance. Our proposed method achieves SoTA or near SoTA performance for eight different NER datasets, proving its generality to various NER subtasks.

\section*{Acknowledgements}
We would like to thank the anonymous reviewers for their insightful comments. The discussion with colleagues in AWS Shanghai AI Lab was quite fruitful. We also thank the developers of fastNLP\footnote{\url{https://github.com/fastnlp/fastNLP}. FastNLP is a natural language processing python package.} and fitlog\footnote{\url{https://github.com/fastnlp/fitlog}. Fitlog is an experiment tracking package.}. We thank Juntao Yu for helpful discussion about dataset processing. This work was supported by the National Key Research and Development Program of China (No. 2020AAA0106700) and National Natural Science Foundation of China (No. 62022027).

\section*{Ethical Considerations}
For the consideration of ethical concerns, we would make detailed description  as following:

(1) All of the experiments are conducted on existing datasets, which are derived from public scientific papers.

(2) We describe the characteristics of the datasets in a specific section. Our analysis is consistent with the results.

(3) Our work does not contain identity characteristics. It does not harm anyone.

(4) Our experiments do not  need a lots of computer resources compared to pre-trained models.

% \section*{Acknowledgments}

% The acknowledgments should go immediately before the references. Do not number the acknowledgments section.
% \textbf{Do not include this section when submitting your paper for review.}

\bibliographystyle{acl_natbib}
\bibliography{anthology,acl2021}

\clearpage
\appendix
\section{Supplemental Material}

\subsection{Hyper-parameters} \label{supply:random_search}
% 这里列出所有的搜索范围。

The detailed hyper-parameter used in different datasets are listed in Table \ref{tb:hyper}. We use the slanted triangular learning rate warmup. All experiments are conducted in the Nvidia Ge-Force RTX-3090 Graphical Card with 24G graphical memory.

\begin{table}[ht]
  \begin{tabular}{l|l}
  \toprule
  Hyper       & Value                                            \\
  \midrule
  Epoch       & 30                                               \\
  Warmup step & 0.01                                             \\
  Learning rate          & [1e-5,2e-5,4e-5] \\
  Batch size  & 16                                               \\
  BART        & Large                                           \\
  $\alpha$    & 0.5                                            \\
  Beam size  & [1, 4]  \\
  \bottomrule
  \end{tabular}
  \caption{Hyper-parameters used for CoNLL2003, OntoNotes, ACE2004, ACE2005, Genia, CADEC, ShARe13, ShARe14.}
  \label{tb:hyper}
\end{table}

\subsection{Beam Search}
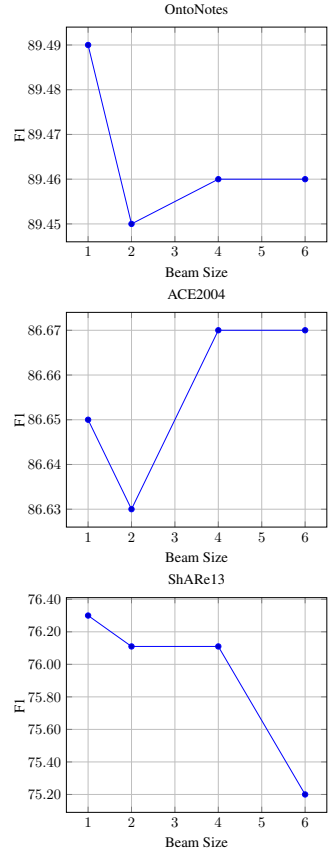
\begin{figure}[htb]
  \centering
  \begin{subfigure}{0.32\textwidth}
  \centering
  \begin{tikzpicture}[scale=0.5]
    \begin{axis}[
      xlabel=Beam Size,
      ylabel=F1,
      grid=major,
      title=OntoNotes
    ]
    \addplot coordinates {
        (1, 89.49)
        (2, 89.45)
        (4, 89.46)
        (6, 89.46)
    };
    \end{axis}
  \end{tikzpicture}
  \end{subfigure}

  \begin{subfigure}{0.32\textwidth}
  \centering
  \begin{tikzpicture}[scale=0.5]
    \begin{axis}[
      xlabel=Beam Size,
      ylabel=F1,
      y tick label style={
        /pgf/number format/.cd,
            fixed,
            fixed zerofill,
            precision=2,
        /tikz/.cd
    },
      grid=major,
      title=ACE2004
    ]
    \addplot coordinates{
      (1, 86.65)
      (2, 86.63)
      (4, 86.67)
      (6, 86.67)
    };
    \end{axis}
  \end{tikzpicture}
  \end{subfigure}
  
  % \quad
  % \vskip 0.15in
  \begin{subfigure}{0.32\textwidth}
  \centering
  \begin{tikzpicture}[scale=0.5]
    \begin{axis}[
      xlabel=Beam Size,
      ylabel=F1,
      y tick label style={
        /pgf/number format/.cd,
            fixed,
            fixed zerofill,
            precision=2,
        /tikz/.cd
    },
      grid=major,
      title=ShARe13
    ]
    \addplot coordinates {
        (1, 76.30)
        (2, 76.11)
        (4, 76.11)
        (6, 75.2)
    };
    % \addlegendentry{Voting}
    \end{axis}
  \end{tikzpicture}
  \end{subfigure}
  % \quad
  \caption{The F1 change curve with the increment of beam size. The beam size has limited effect on the F1 score. }\label{fig:generation_param}
\end{figure}

Since our framework is based on generation, we want to study whether using beam search will increase the performance, results are depicted in Figure \ref{fig:generation_param}, it shows the beam search almost has no effect on the model performance. The litte effect on the F1 value might be caused the the small searching space when generating.

\begin{table*}[!h]
  \centering
  \begin{tabular}{clccc}
  \toprule
  Dataset                     & Model      & Memory & Training Time& Evaluation Time\\
  \midrule
  \multirow{3}{*}{CoNLL-2003} & BERT-MLP   & 7G              & 98s           & 3s              \\
                              & BERT-CRF   & 7G              & 122s          & 5s              \\
                              & Ours(Word)[BART] & 8G              & 115s          & 12s             \\
  \midrule
  \multirow{3}{*}{OntoNotes}  & BERT-MLP   & 7G              & 421s          & 9s              \\
                              & BERT-CRF   & 7G              & 523s          & 13s             \\
                              & Ours(Word)[BART] & 7G              & 493s          & 38s         \\
  \bottomrule    
  \end{tabular}
  \caption{The training memory usage, training time and evaluation time comparison between three models. }
  \label{tb:efficiency}
\end{table*}

\subsection{Efficiency Metrics}
In this section, we compare the memory footprint, training and inference time of our proposed model and BERT-based models. The experiments are conducted on the flat NER datasets, CoNLL-2003 \cite{DBLP:conf/conll/SangM03} and OntoNotes \cite{DBLP:conf/conll/PradhanMXUZ12}. We use the BERT-MLP and BERT-CRF models as our baseline models. BERT-MLP and BERT-CRF are sequence labelling based models. For an input sentence $X=[x_1, ..., x_n]$, both models use BERT \cite{DBLP:conf/naacl/DevlinCLT19} to encode $X$ as follows
\begin{align}
  \mathbf{H} = \mathrm{BERT}(X)
\end{align}
where $\mathbf{H} \in \mathbb{R}^{n \times d}$, $d$ is the hidden state dimension.

Then for the BERT-MLP model, it decodes the tags as follows
\begin{align}
  \mathbf{F} = \mathrm{Softmax}( \mathrm{max}(\mathbf{H}W_b+b_b, 0)W_a + b_a)
\end{align}
where $W_a \in \mathbb{R}^{d \times |T|}$ and $|T|$ is the number of tags, $b_a \in \mathbb{R}^{|T|}$, $W_b \in \mathbb{R}^{d \times d}$, $b_b \in \mathbb{R}^{d}$, $\mathbf{F} \in \mathbb{R}^{n \times |T|}$ is the tag probability distribution. Then we use the negative log likelihood loss. And during the inference, for each token, the tag index with the largest probability is deemed as the prediction. 

For the BERT-CRF model, we use the conditional random fields (CRF) \cite{lafferty2001conditional} to decode tags. We assue the golden label sequence is $Y=[y_1,...,y_n]$, then we use the following equations to get the probability of $Y$
\begin{align}
  \mathbf{M} & = \mathrm{max}(\mathbf{H}W_b+b_b, 0) W_a + b_a \\
  \mathbf{M} & = \mathrm{log\_softmax}(\mathbf{M}) \\
  P(Y|X) & = \frac{\sum_{i=1}^{n}e^{\mathbf{M}[i, y_i]+\mathbf{T}[y_{i-1}, y_i]}}
  {\sum_{y^{\prime}}^{\mathbf{Y}(\mathbf{s})}\sum_{i=1}^{n}e^{\mathbf{M}[i, y^{\prime}_i]+\mathbf{T}[y^{\prime}_{i-1}, y^{\prime}_i]}},
\end{align}
where $\mathbf{M} \in \mathbb{R}^{n \times |T|}$, $\mathbf{Y}(\mathbf{s})$ is all valid label sequences, $\mathbf{T} \in \mathbb{R}^{|T| \times |T|}$ is the transitation matrix, an entry $(i, j)$ in  $\mathbf{T}$ means the transition score from tag $i$ to tag $j$. After getting the $P(Y|X)$, we use negative log likelihood loss to optimize the model. During the inference, the Viterbi Algorithm is used to find the label sequence achieves the highest score. 

We use the BERT-base version and BART-base version to calculate the memory footprint during training, seconds needed to iterate one epoch (one epoch means iterating over all training samples), and seconds needed to evaluate the development set. The batch size is 16 and 48 for training and evaluation, respectively. The comparison is presented in Table \ref{tb:efficiency}. 

During the training phase, we can use the casual mask to make the training of our model in parallel. Therefore, our proposed model can train faster than the BERT-CRF model, which needs sequential computation. While during the evaluating phase, we have to autoregressively generate tokens, which will make the inference slow. Therefore, further work like the usage of a non-autoregressive method can be studied to speed up the decoding \cite{DBLP:conf/iclr/Gu0XLS18}.
\end{document}